\pdfoutput=1

\documentclass[11pt]{article}

\usepackage[]{ACL2023}

\usepackage{times}
\usepackage{latexsym}

\usepackage[T1]{fontenc}

\usepackage[utf8]{inputenc}

\usepackage{microtype}

\usepackage{inconsolata}
\usepackage{booktabs}
\usepackage{graphicx}
\usepackage{multirow}
\usepackage{amsmath}
\usepackage{cleveref}

\usepackage{listings}

\def\footurl#1{\footnote{\url{#1}}}

\title{HausaNLP at SemEval-2025 Task 2: Entity-Aware Fine-tuning vs. Prompt Engineering in Entity-Aware Machine Translation}

\author{Abdulhamid Abubakar$^1$, Hamidatu Abdulkadir$^2$, Ibrahim Rabiu Abdullahi$^3$, \\
{\bf Abubakar Auwal Khalid$^1$, Ahmad Mustapha Wali$^1$, Amina Aminu Umar$^1$, Maryam Bala$^1$,} \\
{\bf Sani Abdullahi Sani$^4$, Ibrahim Said Ahmad$^5$, Shamsuddeen Hassan Muhammad$^6$,} \\
{\bf Idris Abdulmumin$^7$, Vukosi Marivate$^7$} \\
{\small $^1$HausaNLP, $^2$Kaduna State University, $^3$Ahmadu Bello University, $^4$University of the Witwatersrand,} \\
{\small $^5$Northeastern University, $^6$Imperial College, $^7$Data Science for Social Impact, University of Pretoria}\\
\footnotesize\texttt{\textbf{correspondence:} abdulhamid@ab-bkr.com, idris.abdulmumin@up.ac.za}}

\begin{document}
\maketitle

\begin{abstract}
This paper presents our findings for SemEval 2025 Task 2, a shared task on entity-aware machine translation (EA-MT). The goal of this task is to develop translation models that can accurately translate English sentences into target languages, with a particular focus on handling named entities, which often pose challenges for MT systems. The task covers 10 target languages with English as the source. In this paper, we describe the different systems we employed, detail our results, and discuss insights gained from our experiments.
\end{abstract}

\section{Introduction}
The quality of translations produced by machine translation (MT) systems has improved considerably \cite{abdulmumin-etal-2024-correcting}. However, despite these advancements, translations into the target language still contain errors, often due to the challenges associated with translating named entities \cite{rikters-miwa-2024-entity-aware}. Entity-aware machine translation is a type of MT that considers specific entities, such as names, locations, and organizations, to enhance translation accuracy and fluency \cite{conia-etal-2024-towards}. Several approaches have been proposed to improve these models’ ability to translate named entities more effectively, accounting for the need for transliteration in some cases while generating equivalent translations in others.

In this paper, we describe our submissions to SemEval 2025 Task 2: Entity-Aware Machine Translation shared task \cite{ea-mt-benchmark}. Our systems include sequence-to-sequence entity-aware supervised models trained to improve the translation of English sentences into French, German, Spanish, Italian, Japanese, and Arabic. A pre-trained NLLB \cite{nllb} model was fine-tuned for bilingual translation with the provided training data in each of these languages. Furthermore, we investigated the performances of a closed-source Large Language Model (LLM), Gemini \cite{gemini}, on these languages, in addition to Chinese, Korean, Thai, and Turkish. Our results indicate that Gemini, in a zero-shot setup, achieved the best overall performance, with only a few languages showing improvements when examples from the training data were incorporated in few-shot setups.

\section{Related Works}
Several studies have explored approaches to improving named entity (NE) translation in machine translation (MT) systems. \citet{Xie2022-hd} proposed an entity-aware model that employs classifiers in both the encoder and decoder to handle named entities more effectively. Other methods include hierarchical encoders with chunk-based processing \cite{ugawa-etal-2018-neural}, IOB-tagging for improved NE annotation \cite{modrzejewski2020incorporating}, and a decoupled NE handling approach that enhances translation quality without modifying the NMT architecture \cite{mota-etal-2022-fast}. \citet{zeng2023extract} developed an Extract-and-Attend strategy that first identifies and translates named entities separately before incorporating them into the translation process.

Recent advancements also include entity-aware multi-task training on pre-trained models such as T5, which improves NE translation quality and also increases the number of named entities generated in German translations of English texts \cite{rikters-miwa-2024-entity-aware}. \citet{zhiwei2024} introduced the AMFF and CAMFF frameworks, which utilize attention mechanisms to improve named entity recognition (NER) by incorporating multilevel contextual features. \citet{jauhari2024entity} introduced Entity-Aware Techniques (EaT) that integrate semantic parsing to help MT models recognize and accurately translate named entities. \citet{awiszus-etal-2024-charles} evaluated NE translation in speech translation systems, highlighting persistent challenges despite improvements in recall and precision. While these methods improve entity translation, our approach fine-tunes the pre-trained NLLB model using both the provided training data and extracted named-entity translations from Wikidata. Additionally, we evaluate the closed-source Gemini model without fine-tuning, assessing its performance in both zero-shot and few-shot setups. Our work focuses on achieving a balance between overall translation quality and ensuring the accurate translation of critical named entities.

\section{Proposed Approaches}
The team adopted four approaches for this task. The first two approaches included fine-tuning the NLLB pre-trained model in a bilingual setup for \textbf{\texttt{eng}$\rightarrow$\texttt{xxx}} translation. The two other approaches were the use of two different prompt templates to evaluate the performance of Google's Gemini closed-source model. The team combined the traditional fine-tuning technique and prompt-engineering strategies to assess their relative effectiveness in preserving entity integrity during bilingual translation.

\subsection{Supervised Fine-tuning}
This method was employed to train bilingual translation models for six languages, as training data was not available for all target languages. In the first fine-tuning approach, we fine-tuned the base NLLB model using the provided training data. The second approach involved leveraging the SpaCy \cite{honnibal2020spacy} NER framework to extract named entities from the training data and then searching for their equivalent translations on Wikidata \cite{wikidata}. The resulting entity pairs for each source-target language pair were used as additional training data to fine-tune the models. Rather than focusing solely on the named entities provided in the task, we opted to use all entities present in the training data. This approach aimed to improve the models' ability to translate a broader range of entities rather than limiting ourselves to the subset specified in the task. While restricting to the provided subset might have resulted in better performance, we prioritized enhancing the model's accuracy, even at the potential cost of competitiveness. This second approach led to the development of an entity-optimized variant of the NLLB-200 model, refined through targeted fine-tuning using the extracted named entities.

\begin{table*}[t!]
    \small
    \centering
    \begin{tabular}{llrrrrrrrrrr}
    \toprule
       \textbf{metric} & \textbf{method} & \textbf{\texttt{ar}} & \textbf{\texttt{de}} & \textbf{\texttt{es}} & \textbf{\texttt{fr}} & \textbf{\texttt{it}} & \textbf{\texttt{ja}} & \textbf{\texttt{ko}} & \textbf{\texttt{th}} & \textbf{\texttt{tr}} & \textbf{\texttt{zh}} \\
    \midrule
        M-ETA & NLLB+NE & 20.61 & 20.86 & 32.75 & 22.85 & 27.28 & 12.74 & - & - & - & - \\
        & Gemini 0-shot & 32.66 & 38.15 & 47.92 & 38.77 & 40.31 & 35.10 & 34.67 & 18.80 & 40.82 & 8.53 \\
        \cmidrule{2-12}
        & \textbf{Gemini 10-shots+NE} & \textbf{34.17} & \textbf{38.14} & \textbf{48.29} & \textbf{35.32} & \textbf{39.39} & \textbf{34.93} & \textbf{33.75} & \textbf{18.62} & \textbf{41.54} & \textbf{8.09} \\
    \midrule
        COMET & NLLB+NE & 87.98 & 88.42 & 91.26 & 85.07 & 89.52 & 88.86 & - & - & - & - \\
        & Gemini 0-shot & 88.56 & 89.30 & 91.71 & 88.35 & 89.98 & 91.06 & 90.71 & 83.41 & 92.42 & 87.85 \\
        \cmidrule{2-12}
        & \textbf{Gemini 10-shots+NE} & \textbf{89.59} & \textbf{89.86} & \textbf{92.50} & \textbf{88.95} & \textbf{90.64} & \textbf{92.31} & \textbf{91.34} & \textbf{83.97} & \textbf{92.85} & \textbf{88.66} \\
    \midrule
        Overall & NLLB+NE & 33.40 & 33.76 & 48.20 & 36.02 & 41.82 & 22.28 & - & - & - & - \\
        & Gemini 0-shot & 47.72 & 53.46 & 62.95 & 53.89 & 55.68 & 50.67 & 50.17 & 30.68 & 56.63 & 15.55 \\
        \cmidrule{2-12}
        & \textbf{Gemini 10-shots+NE} & \textbf{49.47} & \textbf{53.55} & \textbf{63.45} & \textbf{50.56} & \textbf{54.92} & \textbf{50.68} & \textbf{49.29} & \textbf{30.48} & \textbf{57.40} & \textbf{14.83} \\
    \bottomrule
    \end{tabular}
    \caption{\textbf{M-ETA}, \textbf{COMET}, and the \textbf{Overall} scores of the evaluated approaches. The scores in \textbf{bold} font indicate our final system submission, representing our ranked system according to the task instructions.}
    \label{tab:model_performances}
\end{table*}

\subsection{Prompt Engineering}
Complementing these fine-tuning methods, two prompt-based strategies were implemented. The zero-shot approach utilized minimalist templates instructing the model to "preserve entity integrity" during translations. The few-shot variant extended this with 10 curated demonstration pairs from the training data in the task repository. For target languages without training data, the template provided no examples. The examples followed a structured template showing entity preservation by providing the Wikidata-id of the entity as present in the training dataset.

\section{Experiments}
\subsection{Dataset and pre-processing}
The dataset used to fine-tune the NLLB-200 model was made available as part of the shared task \cite{conia-etal-2024-towards}, with the number of data points in the different splits detailed in \Cref{tab:data_stats}. This repository included training data for English-to-Arabic, German, French, Spanish, Italian, and Japanese translations, which were also used for the few-shot prompting approach. Additionally, the validation and test data for all 10 languages are provided, with the test data being unlabelled.

We also extracted the NEs from Wikidata\footurl{https://www.wikidata.org/wiki/Wikidata:Main_Page}, resulting in 4,587 unique entities and their translations (where available); see \Cref{tab:entity_stats}. We used these entities when fine-tuning the NLLB model and as part of the few-shot prompting. Training the NLLB model required tokenizing the training data; this was achieved using the NLLB tokenizer. The prompt engineering approach did not require any data preprocessing.

\subsection{Models and environment setup}
The fine-tuning approach involved training the NLLB-200 model, specifically the distilled 600M\footnote{https://huggingface.co/facebook/nllb-200-distilled-600M} parameter variant. The fine-tuning was conducted using the default Hugging Face hyperparameter setup: a batch size of 32, sequence lengths of 128 for both source and target, a generation beam search width of 5, a dropout rate of 0.1, and a training duration of 10 epochs. Early stopping was applied if there was no improvement in the model’s performance after two consecutive epochs.

For the prompt engineering method, we utilized Langchain's \cite{langchain} ChatPromptTemplate\footurl{https://python.langchain.com/api_reference/core/prompts/langchain_core.prompts.chat.ChatPromptTemplate.html} and ChatGoogleGenerativeAI\footurl{https://python.langchain.com/api_reference/google_genai/chat_models/langchain_google_genai.chat_models.ChatGoogleGenerativeAI.html} modules to evaluate the performance of Gemini Flash 1.5\footurl{https://ai.google.dev/gemini-api/docs/models/gemini\#gemini-1.5-flash}. We provide the prompt templates that were used in Templates \ref{prompt:template1} and \ref{prompt:template2}. The results were saved in JSON format as required from the task submission description.

\subsection{Evaluation}
We used the metrics provided for the shared task to evaluate our systems. These metrics are COMET \cite{rei-etal-2020-comet} and Manual Entity Translation Accuracy (M-ETA; \citealt{conia-etal-2024-towards}). COMET is a machine translation evaluation metric that leverages a pre-trained model to generate quality scores by comparing system outputs to human translations. M-ETA assesses the accuracy of entity translations in machine translation by computing the proportion of correctly translated entities against a gold standard. Untranslated source texts are scored 0. The overall score, \Cref{eq:1}, is computed as the harmonic mean (F1 score) of the COMET and M-ETA metrics, ensuring that systems are rewarded for balanced performance across both rather than excelling in only one.
\begin{equation}
    \text{Overall Score} = 2 \times \dfrac{\text{COMET} \times \text{M-ETA}}{\text{COMET} + \text{M-ETA}}\label{eq:1}
\end{equation}

\section{Results}
\Cref{tab:model_performances} below gives a summary of the models' performances across the proposed approaches. The NLLB column contains only results for languages that have training data.

The results indicate that while all 3 approaches achieved similar COMET scores at sentence level, the Gemini model, both in it's 0- and few-shot settings, performed better at translating the named entities in the source text. This can be observed by the higher M-ETA scores obtained by Gemini compared to NLLB. In almost all the target languages where we have results, Gemini has almost twice the M-ETA scores of NLLB. At the language level, English-to-Spanish translation achieved the best performance across all the evaluated approaches on all the evaluation metrics. This is in contrast to the Chinese language translation from English, with a paltry overall score of 14.83.

It is important to highlight the performance disparity between European and Asian languages. European languages, including Spanish, Italian, French, and German, consistently achieved higher scores across all methods, except for Japanese. Italian, for instance, achieved scores of 55.68 and 54.92 with Gemini’s approaches, while NLLB managed 22.28. In contrast, Asian languages presented lower scores, with Chinese recording the lowest scores (15.55 and 14.83) among all languages tested. Japanese, while performing moderately well with Gemini (50.57 and 50.68), showed lower scores compared to European languages in NLLB but higher than the other Asian language: Arabic. Overall, Spanish and Italian achieved better results than the others in NLLB.

As highlighted earlier, not all the target languages had training data, so this affected the testing of these languages with the NLLB model, thus resulting in no scores for Chinese, Korean, Thai, and Turkish. However, an interesting finding was Turkish’s performance with Gemini (56.63 and 57.40) despite the absence of few-shot examples, yet it's performance was second to only Spanish.

A comparison of the two Gemini implementations revealed minimal differences between the zero-shot and few-shot approaches. This suggests that elaborate few-shot prompting may not be necessary to achieve optimal results in entity translation tasks.

\section{Conclusion}
Prompt engineering for large language models like Gemini proves effective for entity-aware machine translation. Comparative studies with fine-tuning the NLLB-200 model show a consistent performance advantage for Gemini’s zero-shot and few-shot prompting across target languages, highlighting its ability to preserve entity integrity in translation. While our results reveal nuances in performance across language families, with European languages exhibiting stronger overall scores and an intriguing performance from Turkish, the overarching trend favors prompt-based methodologies.

\section{Limitations}
Our evaluation was constrained by the lack of training data for Chinese, Korean, Thai, and Turkish in the task repository, preventing us from fine-tuning NLLB models for these languages. Additionally, due to limited computational resources, we were only able to fine-tune the distilled 600M-parameter variant of NLLB-200, rather than a larger model that could potentially yield better results.

\section*{Ethics Statement}
This work followed the guidelines provided in SemEval 2025.

\section*{Acknowledgements}
The authors thank HausaNLP, particularly Dr. Idris Abdulmumin, Dr. Shamsudden Hassan Muhammad, and Dr. Ibrahim Said Ahmad, for their invaluable mentorship throughout the process.

\bibliography{anthology,mybib}
\bibliographystyle{acl_natbib}

\appendix

\section{Appendix}
We provide the dataset and extracted named entity statistics and also the prompt templates used in the experiments.

\begin{table}[t!]
    \centering
    \begin{tabular}{lrrr}
    \toprule
        language & train & validation & test \\
    \midrule
        \textbf{Arabic (\texttt{ar})} & 7,220 & 722 & 4,547 \\
        \textbf{German (\texttt{de})} & 4,087 & 731 & 5,876 \\
        \textbf{Spanish (\texttt{es})} & 5,160 & 739 & 5,338 \\
        \textbf{French (\texttt{fr})} & 5,531 & 724 & 5,465 \\
        \textbf{Italian (\texttt{it})} & 3,739 & 730 & 5,098 \\
        \textbf{Japanese (\texttt{ja})} & 7,225 & 723 & 5,108 \\
        \textbf{Korean (\texttt{ko})} & - & 745 & 5,082 \\
        \textbf{Thai (\texttt{th})} & - & 710 & 3,447 \\
        \textbf{Turkish (\texttt{tr})} & - & 732 & 4,473 \\
        \textbf{Chinese (\texttt{zh})} & - & 722 & 5,182 \\
    \bottomrule
    \end{tabular}
    \caption{Shared task data statistics.}
    \label{tab:data_stats}
\end{table}

\onecolumn
\small
\begin{lstlisting}[caption={Template 1}, label={prompt:template1}]
PromptTemplate(
    input_variables=["sentence", "tgt", "ne", "examples"],
    template="""Instruction:
    Translate the following text from english to {tgt}, ensuring that all 
    named-entities are accurately translated with no additional explanations. Use 
    the provided translation examples and translated named-entities (if provided) 
    for consistency. Do not send the English text back in the response, generate 
    only the translation and nothing more.
    Named entities:
    {ne}
    Examples:
    {examples}
    Now generate the {tgt} translation of the following english text: {sentence}"""
)
\end{lstlisting}

\begin{lstlisting}[caption={Template 2}, label={prompt:template2}]
PromptTemplate(
    input_variables=["sentence", "tgt"],
    template="""Instruction:
    Translate the following text from english to {tgt}, ensuring that all named-
    entities are accurately translated with no additional explanations. Do not send 
    the English text back in the response, generate only the translation and nothing 
    more.
    Now generate the {tgt} translation of the following english text: {sentence}"""
)
\end{lstlisting}

\begin{table*}[t!]
    \small
    \centering
    \resizebox{\textwidth}{!}{
    \begin{tabular}{lrrrrrrrrrrrr}
    \toprule
    & \multicolumn{11}{c}{\textbf{count}} \\ \cmidrule{2-12}
    \textbf{Entity type} & \textbf{all} & \textbf{\texttt{ar}} & \textbf{\texttt{de}} & \textbf{\texttt{es}} & \textbf{\texttt{fr}} & \textbf{\texttt{it}} & \textbf{\texttt{ja}} & \textbf{\texttt{ko}} & \textbf{\texttt{th}} & \textbf{\texttt{tr}} & \textbf{\texttt{zh}} \\
    \midrule
    PERSON & 1,507 & 908 & 1,083 & 1,081 & 1,114 & 1,069 & 1,021 & 941 & 673 & 933 & 1,014 \\
    ORG & 1,082 & 648 & 785 & 788 & 788 & 767 & 745 & 680 & 470 & 660 & 734 \\
    GPE & 522 & 291 & 360 & 363 & 369 & 351 & 333 & 307 & 201 & 292 & 330 \\
    DATE & 379 & 221 & 264 & 269 & 268 & 263 & 246 & 229 & 160 & 231 & 248 \\
    WORK\_OF\_ART & 282 & 171 & 209 & 205 & 214 & 207 & 194 & 175 & 122 & 178 & 193 \\
    EVENT & 187 & 105 & 135 & 137 & 138 & 134 & 121 & 105 & 73 & 108 & 120 \\
    LOC & 183 & 102 & 127 & 128 & 128 & 125 & 118 & 101 & 69 & 105 & 121 \\
    NORP & 169 & 103 & 124 & 124 & 125 & 122 & 116 & 105 & 67 & 98 & 116 \\
    FAC & 135 & 78 & 94 & 95 & 96 & 94 & 88 & 80 & 43 & 80 & 87 \\
    PRODUCT & 51 & 37 & 42 & 43 & 43 & 42 & 40 & 37 & 28 & 37 & 39 \\
    LAW & 31 & 21 & 23 & 24 & 23 & 23 & 22 & 21 & 16 & 22 & 22 \\
    QUANTITY & 28 & 14 & 19 & 19 & 20 & 20 & 18 & 14 & 10 & 13 & 18 \\
    MONEY & 15 & 5 & 8 & 8 & 8 & 8 & 6 & 5 & 5 & 6 & 7 \\
    TIME & 10 & 5 & 7 & 7 & 7 & 7 & 7 & 4 & 3 & 5 & 6 \\
    PERCENT & 4 & 1 & 2 & 3 & 3 & 2 & 3 & 2 & 1 & 1 & 2 \\
    LANGUAGE & 2 & 1 & 2 & 2 & 2 & 2 & 1 & 1 & 1 & 1 & 1 \\
    \bottomrule
    \end{tabular}
    }
    \caption{Entities extracted from training data and their translations obtained from Wikidata.}
    \label{tab:entity_stats}
\end{table*}

\end{document}